\documentclass[letterpaper, 10 pt, conference]{ieeeconf}  

\IEEEoverridecommandlockouts                              

\usepackage{amsmath} 
\usepackage{amssymb}  
\usepackage{dsfont}
\usepackage[utf8x]{inputenc}
\usepackage{graphicx}
\usepackage{atbegshi,picture}

\newcommand{\RR}{\mathds{R}} 
\renewcommand{\vec}[1]{{\boldsymbol{{#1}}}} 
\newcommand{\mat}[1]{{\boldsymbol{{#1}}}} 

\title{\LARGE \bf
HemCNN: Deep Learning enables decoding of \\ fNIRS cortical signals in hand grip motor tasks
}

\AtBeginShipoutNext{\AtBeginShipoutUpperLeft{%
  \put(\dimexpr\paperwidth-1cm\relax,-1.5cm){\makebox[0pt][r]{Accepted manuscript IEEE/EMBS Neural Engineering (NER) 2021}}%
}}

\author{Pablo Ortega$^{1}$ and Aldo Faisal$^{1,2}$
\thanks{Brain \& Behaviour Lab, $^{1}$Department of Computing,
        $^{2}$Department of Bioengineering, Imperial College London,  
        London SW7 2AZ, UK; Correspondence: aldo.faisal@imperial.ac.uk}%
}

\begin{document}

\maketitle
\thispagestyle{empty}
\pagestyle{empty}

\begin{abstract}
We solve the fNIRS left/right hand force decoding problem using a data-driven approach by using a convolutional neural network architecture, the HemCNN.  We test HemCNN's decoding capabilities to decode in a streaming way the hand, left or right, from fNIRS data. 
HemCNN learned to detect which hand executed a grasp at a naturalistic hand action speed of  $~1\,$Hz, outperforming standard methods.  
Since HemCNN does not require baseline correction and the convolution operation is invariant to time translations, our method can help to unlock fNIRS for a variety of real-time tasks. 
Mobile brain imaging and mobile brain machine interfacing can benefit from this to develop real-world neuroscience and practical human neural interfacing based on BOLD-like signals for the evaluation, assistance and rehabilitation of force generation, such as fusion of fNIRS with EEG signals. 
\end{abstract}

\section{INTRODUCTION}
Non-invasive neuroimaging techniques provide an easy way to study the human cortex in a variety of tasks and provide opportunities for decoding mental activity. 
Especially, in sensorimotor tasks that reflect on ecologically valid real-world tasks, these methods can help us to evaluate, rehabilitate, replace or assist the motor function after brain damage \cite{ENRIQUEZGEPPERT20131}. 

We suggest that model-free and non-linear methods can be more suitable for capturing relevant hemodynamic response (HR) variability from functional near-infrared spectroscopy (fNIRS) during force generation.
In particular, differences in force generation tasks may explain why some force-tasks evoke spatially-resolved HR that reveal hemispherical lateralisation differences, a result often sought in motor control.
For example, finger pinches in \cite{Nambu2009a} or finger tapping in \cite{trakoolwilaiwan2017convolutional} achieve lateralisation results while other work using hand grips do not  \cite{shibuya2008quantification,Derosiere2014a,wriessnegger2017force}.
These works have in common the use of linear techniques and feature engineering in their methodology.
For a given task, linear techniques first require to compute an expected HR and then regress out all the variability in the fNIRS measurements that does not fit the expected HR.
These linear techniques, like general linear models (GLM), have proven very useful in the reduction of these sources of noise \cite{friston1994statistical,ye2009nirs}.
Nonetheless, they require previous knowledge of the relationship between the task under study and the HR in order to compute accurate HR predictions that capture the variability in the task.
If the relationship of the HR with the task is unknown or inaccurate, the variability in the HR might end up being discarded as noise instead of being explained by the underlying neuronal process \cite{faisal2008noise}.
This limitation can be particularly relevant in the study of motor control where it has been observed that the speed \cite{kuboyama2005relationship} and the intensity \cite{shibuya2008quantification,Nambu2009a,Derosiere2014a} of a motor execution modulate the HR yet the analytical relationships are unknown.

To advance the neuroimaging capabilities of fNIRS in human force decoding, we set out to demonstrate how Deep Learning (DL) can be used to improve the decoding and interpretation of the HR. 
Convolutional neural networks (CNN) have been introduced to decode brain signals  \cite{walker2015deep}
and since then applied to fNIRS in \cite{trakoolwilaiwan2017convolutional} to detect motor activity. 
However, they did not study force generation nor developed an architecture that reflects our understanding of cortical motor neuroscience.
We present HemCNN, a DL approach based on CNN, and use it to decode spatio-temporally specific cortical activations during force generation.  
We hypothesise that our HemCNN architecture is better than linear methods at processing highly variable HR evoked during fast hand-grips.

\section{METHODS AND MATERIALS}

\begin{figure}[b]
    \centering
    \includegraphics[width=0.45\textwidth]{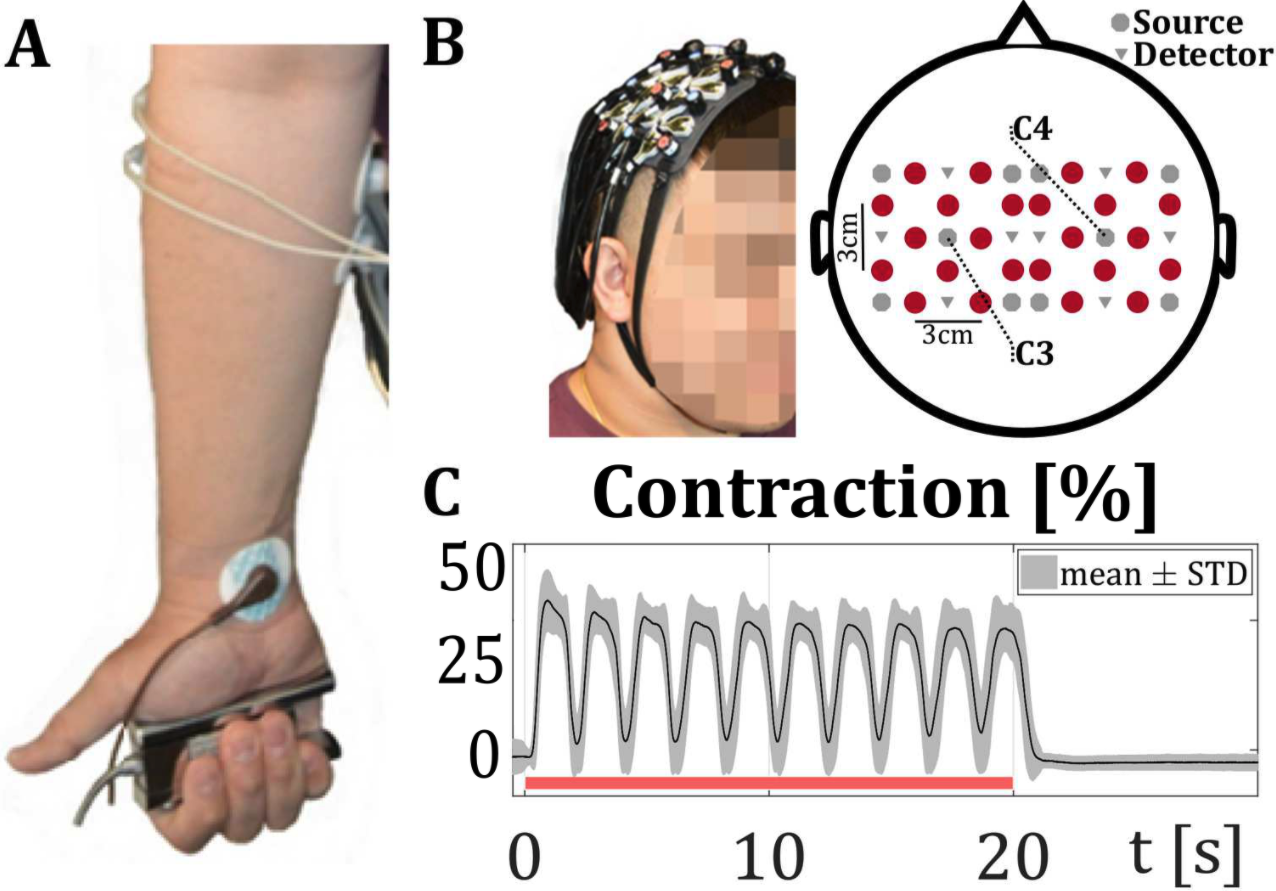}
    \caption{(A) Grip force pose, (B) fNIRS recording arrangement, (C) averaged grip signals.}
    \label{fig:setup}
\end{figure}

\subsection{Protocol and task}

We use a unimanual hand grip task and simultaneously record the fNIRS signals during its realisation in $12$ participants.
The hand force grip task consists of periodic unimanual grips at an approximate rate of $1\,$Hz.
The contraction target was set in the $25-50\%\,$ maximum voluntary contraction (MVC) range.
The grip force task was measured with a grip force transducer (Fig. \ref{fig:setup}A, PowerLab 4/25T, ADInstruments, Castle Hill, Australia) which was used to give visual feedback on contraction.
Trials lasted $21\,$s followed by a randomised resting period uniformly distributed between $15$ and $21\,$seconds, so as to avoid phasic constructive interference of any systemic artefacts in the brain signals (Fig. \ref{fig:setup}C).
Each trial was performed with either hand at different times.
Participants performed a $10$ trials with each hand.
All participants were healthy and right-handed. 
Handedness was confirmed with the Edinburgh inventory \cite{oldfield1971assessment} for all participants.
Imperial College Research Ethics Committee approved all procedures and all participants gave their written informed consent.  

Brain fNIRS signals were recorded using an NIRScout system (NIRx Medizintechnik GmbH, Berlin). 
We used a total of 24 channels (10 sources and 8 detectors) sampling at $12.6\,$Hz.
These channels were symmetrically laid around C3 and C4 positions according to the International 10-20 system leaving an inter-optode distance of $3\,$cm covering the sensorimotor cortex.
Two wavelengths ($wl_1=760\,$nm, $wl_2=850\,$nm) continuous wave near-infrared spectroscopy was used to obtain optical absorption densities that were transformed to oxy-hemoglobin (HbO) and deoxy-hemoglobin  concentrations (HbR) using the modified Beer-Lambert Law \cite{cope1988methods}.
HbO and HbR concentrations ($\mbox{Hb}_C$ to refer to either concentration) were down-pass filtered below $0.25\,$Hz, linearly detrended over a $10\,$s period and downsampled to $2\,$Hz.  

Finally, breathing was recorded as the chest diametrical expansion measured by a transducer ($3\,$Hz sampling rate, PowerLab 4/25T, ADInstruments).
This additional signal was used to control for possible hand-use induced systemic artefacts that could contaminate the fNIRS signals. 

The data used for this work has been made publicly available in the HYGRiP dataset \cite{ortega2020hygrip}.

\subsection{Baseline classification methods}

The following conventional methods are used to compare and assess the performance of HemCNN.
We extract (1) general linear model (GLM) features, (2) lateralisation indices (LI) and (3) concentration changes from the fNIRS signal.
Each feature set is used to train an independent tree classifier following the same leave-one-subject out strategy than HemCNN.

In particular, GLM with canonical HRs (gamma functions) is the current standard in fNIRS analysis for denoising \cite{ye2009nirs}.
After GLM, each channel of each $20\,$s trial segment is represented by a $\vec v \in \RR^5$. 
Each element in $\vec v$ represents approximately $\frac{1}{5}$ of the segment's time-series.
Therefore, the five features represent a chronologically ordered abstract representation of the time-series.
The total dimension is $240$ ($5 \times 48$ HbO and HbR channels).
We additionally used PCA to reduce this dimension in two hierarchical steps (GLM-hPCA).
First across HbO$/$HbR pairs and, second, per hemisphere. 
This process reduced the dimensionality to $10$ ($5 \times 2$ hemispheres).

LI are computed per each pair of symmetric channels leading to a 24-dimensional feature per example.
$LI>0$ indicates right hemisphere dominance and $LI<0$ left hemisphere dominance.

\subsection{HemCNN design and training}

CNN optimise convolutional filters (CF) to extract relevant features for a machine learning task, in our case, classification.
We designed the HemCNN architecture and its training to enhance hemispherical differences between HRs corresponding to the left or the right hand generated force.

Specifically, the classifier corresponds to a mapping 
$\mat f:\RR^{\,\mbox{\footnotesize{nCh}}\,\times\,\mbox{\footnotesize{nSamps}}}\longrightarrow\RR^{\,2}$.
Each dimension in the output space represents the logarithmic probability of an fNIRS example ($X \subset \RR^{\,\mbox{\footnotesize{nCh}}\,\times\,\mbox{\footnotesize{nSamps}}}$) belonging to the left or right-hand generated force,
$ f(\mat X) = \log \left( p(\,\mbox{hand}\,|\,\mat X) \right)$.

Constraints are introduced in the architecture in the form of CF parameters that only allow interaction in the Hb, time and channel dimensions of the fNIRS signals. 
In particular, each HemCNN output value represents a hand and corresponds to the ipsilateral hemisphere. 
Classification decisions are exclusively based on hemispherical differences. 
Furthermore, each hand activity can be easily traced back through the layers to each hemisphere.

\begin{figure}[b]
    \centering
    \vspace{0.3cm}
    \includegraphics[width=0.45\textwidth]{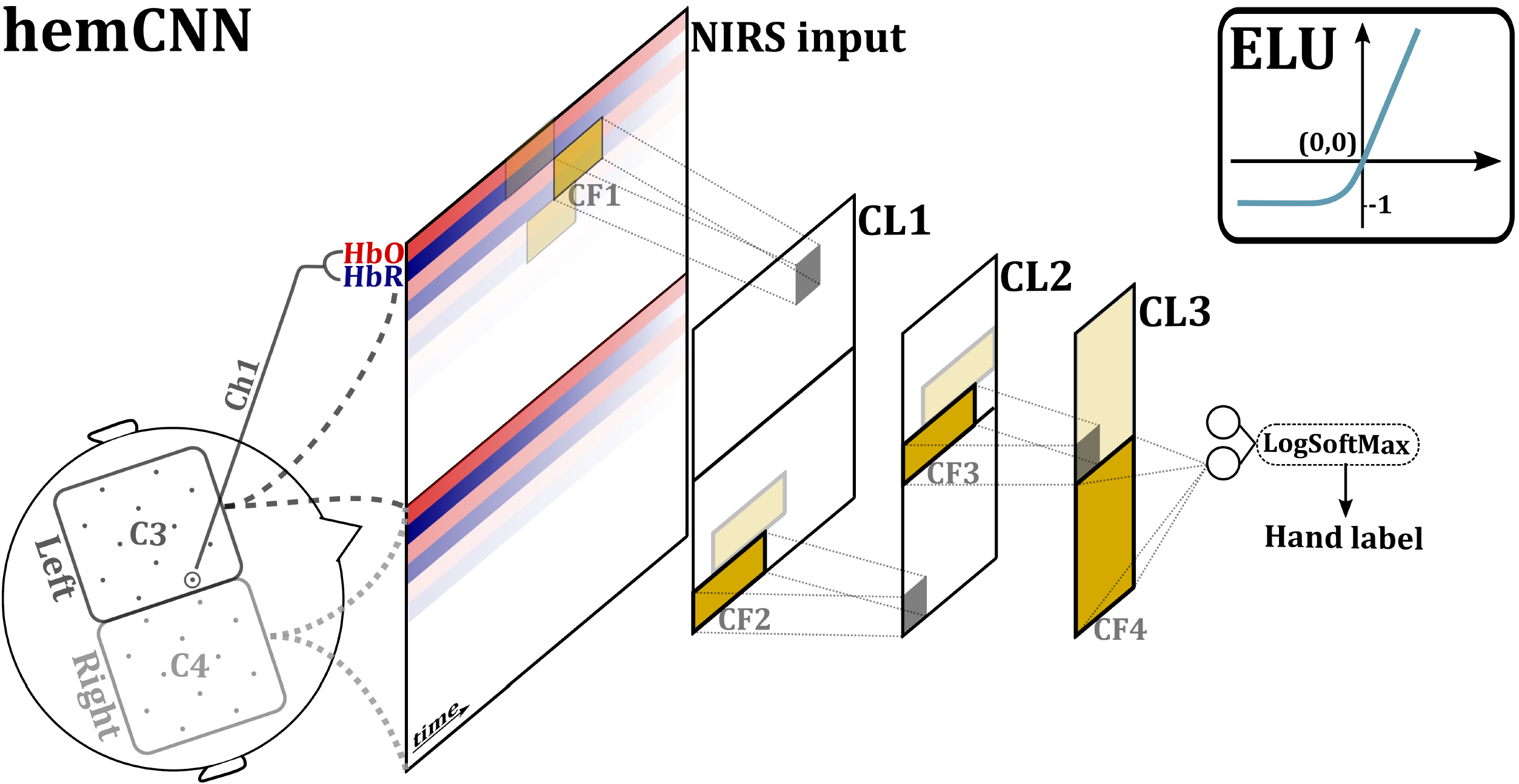}
    \caption{
    HemCNN architecture. 
    From left to right: Channels in the input are arranged per row as HbO$/$HbR consecutive pairs. 
    The upper half of the rows corresponding to the left hemisphere. 
    The bottom half corresponds to the right hemisphere.
    Columns correspond to time steps. 
    Each convolutional filter (CF) renders a deeper convolutional layer activation (CL) after convolution.
    Hand activities can be unequivocally traced back to throughout all layers.
    }
    \label{fig:scheme_arch}
\end{figure}

Fig. \ref{fig:scheme_arch} shows our HemCNN architecture design with the fNIRS input, 4 CF leading to 3 corresponding convolutional layers (CL) and one output layer with two output units representing both hands. 
The CF convolve their respective input space leading to an output that is reduced in dimension depending on the filter size. 
We tailor the shapes of the filters and their stride to introduce constraints in the dimensions in which the CNN will be able to find relationships.
Rows correspond to channels and columns to sampling time-steps.
In a CF, the kernel, $\mat W \subset \RR^{\,m'\,\times\,n'} $, is the set of weights that the input is convolved with, and the stride is the number of steps the kernel shifts per row and column during the convolution. 
In Table \ref{tab:HemCNN} HemCNN kernel sizes and strides are shown.
In particular, our HemCNN implementation preserves the brain hemispherical origin of information throughout all the layers.

We use the cross-entropy loss and the Adam optimiser \cite{kingma2014adam} to train the network with a $10^{-3}$ weight decay.
The training batches contain $10$ left/right balanced shuffled examples and the learning rate starts at $0.03$.
Training is iterated for $15$ epochs (i.e. $15$ times over the augmented training-set) with the learning rate decaying at a $0.9$ rate per epoch.

During training, the drop-out technique is used to regularise the training \cite{srivastava2014dropout}.
Zeroing masks are applied to the fNIRS inputs affecting one randomly selected brain hemisphere with equal probability.
This allows us to consistently obtain models that better reflect the contrast between hemispherical activities for each hand.

The fNIRS time-series are z-scored per trial using the mean and standard deviation of the whole channel set per Hb type.
In contrast to the baseline conventional approaches, the HRs are not baseline corrected.
Finally, the training set is also $5-$fold augmented by 
(1) randomly cropping $19\,$s time windows from an initial $0-35\,$s trial window starting at the go-cue and 
(2) by multiplying each trial by a random number drawn from a uniform distribution between $0.5$ and $1.0$.

\begin{table}[b]
    \caption{Kernel shape (rows $\times$ columns) and stride (rows $\times$ columns)}
    \centering
    \begin{tabular}{ccccc}
    \textbf{HemCNN}         & CF1          & CF2        & CF3        & CF4         \\ \hline
    kernel ($m \times n$)   & $2\times10$  & $1\times5$ & $1\times5$ & $12\times5$ \\ 
    stride ($m \times n$)   & $2\times1$   & $1\times2$ & $1\times2$ & $12\times5$ \\ \hline
    Total                   & 94&&&\\
    \end{tabular}
    \label{tab:HemCNN}
\end{table}

We test HemCNN following a leave-\-one-\-subject-\-out (LOSO) testing approach.
LOSO is a particularly demanding generalisation condition that requires the decoding to operate on subjects who rarely share the same anatomy and sensor placements with the training subjects \cite{xiloyannis2015gaussian}.
For each subject $20$ balanced left/right-hand trials are used as training-set. 
For each left-out subject, $55\%$ of data from the remaining subjects is used for training and $45\%$ for validation.
Training and validation data are split randomly every run and $10$ runs per left-out subject are executed delivering $10$ different models per left-out subject.  
Each model has a different random initialisation and trains for $15$ epochs.
Finally, every model is trained with $605$ augmented examples.
Training time was under $1\,$min/subject in a GeForce RTX2080.
The model with highest validation accuracy is selected and the $20$ test trials of the left-out subject are used to produce all results per subject.

\vspace{-0.25cm}
\section{RESULTS}
To evaluate the performance of HemCNN resolving the HR in fNIRS we used the data of $12$ participants from HYGRiP performing a left/right hand-grip task.
The fNIRS signals are used in this study.
HemCNN is trained to detect what hand (left or right) is being used from these HRs as in a BCI decoding paradigm. 

\begin{figure}[b]
    \centering
    \includegraphics[width=0.47\textwidth]{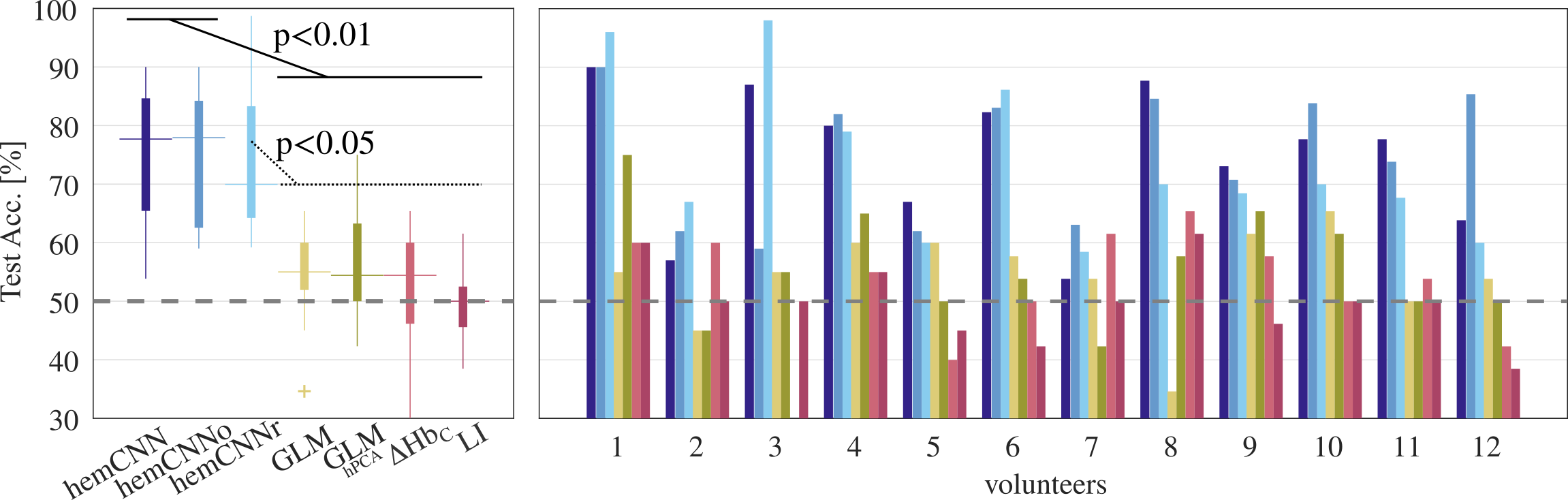}
    \caption{
    Leave-one-subject-out test accuracy median across subjects (left) and per subject (right) for the best validation model out of $10$ trained models.
    Cold colours denote CNN architectures: HemCNN with HbO (HemCNNo), HbR (HemCNNr) or otherwise both inputs (HemCNN). 
    Warm colours denote standard methods to extract features from HbO and HbR which are classified using a simple tree: general linear model (GLM) features, GLM with hierarchical PCA features (GLM-hPCA), $\Delta \mbox{Hb}$ features and LI features. 
    }
    \label{fig:classification}
\end{figure}

We first measure the ability to distinguish between HR evoked by the left or the right hand grips using the classification accuracy of the decoding methods.
Figure \ref{fig:classification} presents the hand test classification accuracy for HemCNN, GLM, and GLM-hPCA along with that of the traditional $\Delta\mbox{Hb}$ and LI features.
Except HemCNN, the remaining methods use a tree classifier with the respective features: GLM, GLM-hPCA, $\Delta\mbox{Hb}$ or LI.
The figure shows that HemCNN methods outperform GLM, GLM-hPCA, $\Delta\mbox{Hb}$ and LI approaches (Kruskal-Wallis test, Tukey correction for pairwise comparison, $\alpha=0.05$).
The use of both HbO and HbR compared to only using one of them does not lead to significant differences although the absence of HbO in HemCNNr leads to a drop in accuracy (median accuracies, HemCNN $77.7\%$, HemCNNo $77.9\%$ and HemCNNr $69.3\%$, $p>0.9$).
Non HemCNN methods, except LI, show similar performances around $54.4-55.0\%$ median accuracies. 
In particular, the addition of hierarchical PCA in GLM leads to the highest performance in non HemCNN methods (subject 1, accuracy $75\%$).
LI performs with a random median accuracy level ($50\%$).
The low performance of non HemCNN approaches show that the classification rules they learn do not generalise well to unseen subjects.
In contrast, HemCNN, provides higher accuracies with stable generalisation across subjects.
This suggests that our proposed method learns HR features that are physiologically shared across subjects.
Moreover, our HemCNN design is particularly suited for interpretation.
On the one hand, it allows the activity of each hand to be traced back through each layer to its hemispherical hemodynamic activity.
On the other hand, each CF has an unambiguous signal processing role inside the network. 
For example, CF1 convolves HbO$/$HbR pairs and CF4 convolve cross channel activities.
Note that a standard CNN with 3 fully connected layers and 4 convolutional filters per convolutional layer achieved a median accuracy of $80.3\%$ which was not significantly different to that of HemCNN ($77.7\%$, Kruskal-Wallis, Tukey corrected test, $\alpha=0.05$).
However, compared to HemCNN, such architecture would not allow to unequivocally trace back the activity of each output.
Instead, these activities would be sparsely distributed across different filters and fully connected units and would not guarantee that the classification is based, exclusively, in interhemispheric differences.
Thus we did not include the standard CNN in the full analysis.

Finally, when the breathing signal was used with an architecture similar to HemCNN to decode the used hand, the resulting accuracy was at chance level ($50\%$).
The lack of accuracy of the breathing rate to decode what hand was used shows that this systemic artefact does not carry hand related information during the task and is unlikely informing the fNIRS based decoding of HemCNN.

\vspace{-0.2cm}
\section{DISCUSSION}

We presented a novel DL approach, HemCNN, to perform decoding and neural data analysis.
Our left/right hand grip task evokes highly variable and non-stationary hemodynamic responses \cite{ortega2020hygrip}.  
Our HemCNN method outperformed currently used methods, specially GLM, at decoding the hand used.
In particular, HemCNN detected the hand used in our task based only in hemispherical differences.

Previous studies using a similar task could not resolve these differences \cite{Derosiere2014a} using linear models or feature engineering or took more than $30\,$s of stimulation to detect them \cite{shibuya2008quantification}. 
In contrast, we do not to make any assumptions on the HR features present during force generation and developed a CNN for fNIRS time-series decoding called HemCNN. 
In particular, HemCNN performs significantly better than ($\sim 78\%$, $p<0.01$, Fig. \ref{fig:classification}) GLM features \cite{ye2009nirs} and other conventional engineered features (Lateralisation indices and hemoglobin concentration changes).

Our analysis demonstrates that differences in fNIRS signals exist across hands but require appropriate feature learning to be detected. 
Furthermore, current understanding tend towards both hemispheres acting together to produce complex unimanual and bimanual movements \cite{haar2017effector,Ames2019}. We show here that lateralised temporal brain activity patterns can be exploited for lateralised and fast hand grips decoding in fNIRS.
We use a light-weight data driven method that simultaneously optimises the localisation of hemodynamic response transients, their time alignment, and their channel relationship by means of the introduction of non-linearities.

Our results for fNIRS force decoding brings us closer to the range of ability of machine learning based force decoding from peripheral sensory signals
\cite{gavriel2014comparison,fara2014prediction}.

\vspace{-0.1cm}
\section{CONCLUSION}
\vspace{-0.1cm}
In this work, we introduced an interpretable CNN based method to temporally and spatially resolve highly variable cortical hemodynamic responses during a demanding motor task.
In particular, HemCNN can be directly applied in real-time analysis since: the convolutional operator is invariant to time translations, the differences are found across time-samples and channels, and there is no strong preprocessing other than the z-scoring of the fNIRS example. 
This makes it appropriate not only for offline neuroimaging but also for BCI paradigms.





\vspace{-0.25cm}
\section*{ACKNOWLEDGMENTS}
\vspace{-0.1cm}
We thank Tong Zhao for support in data recording, and EPSRC for financial through HiPEDS CDT (EP/L016796/1) and an  EPSRC capital equipment grant.


\bibliographystyle{ieeetrans}
\bibliography{bibliography.bib}

\begin{thebibliography}{10}
\providecommand{\url}[1]{#1}
\csname url@samestyle\endcsname
\providecommand{\newblock}{\relax}
\providecommand{\bibinfo}[2]{#2}
\providecommand{\BIBentrySTDinterwordspacing}{\spaceskip=0pt\relax}
\providecommand{\BIBentryALTinterwordstretchfactor}{4}
\providecommand{\BIBentryALTinterwordspacing}{\spaceskip=\fontdimen2\font plus
\BIBentryALTinterwordstretchfactor\fontdimen3\font minus
  \fontdimen4\font\relax}
\providecommand{\BIBforeignlanguage}[2]{{%
\expandafter\ifx\csname l@#1\endcsname\relax
\typeout{** WARNING: IEEEtran.bst: No hyphenation pattern has been}%
\typeout{** loaded for the language `#1'. Using the pattern for}%
\typeout{** the default language instead.}%
\else
\language=\csname l@#1\endcsname
\fi
#2}}
\providecommand{\BIBdecl}{\relax}
\BIBdecl

\bibitem{ENRIQUEZGEPPERT20131}
S.~Enriquez-Geppert, R.~J. Huster, and C.~S. Herrmann, ``Boosting brain
  functions: Improving executive functions with behavioral training,
  neurostimulation, and neurofeedback,'' \emph{Intl. J. Psychophysiology},
  vol.~88, no.~1, pp. 1 -- 16, 2013.

\bibitem{Nambu2009a}
I.~Nambu, R.~Osu, M.~aki Sato, S.~Ando, M.~Kawato, and E.~Naito,
  ``{Single-trial reconstruction of finger-pinch forces from human
  motor-cortical activation measured by near-infrared spectroscopy (NIRS)},''
  \emph{NeuroImage}, vol.~47, no.~2, pp. 628--637, 2009.

\bibitem{trakoolwilaiwan2017convolutional}
T.~Trakoolwilaiwan, B.~Behboodi, J.~Lee, K.~Kim, and J.-W. Choi,
  ``Convolutional neural network for high-accuracy functional near-infrared
  spectroscopy in a brain--computer interface: three-class classification of
  rest, right-, and left-hand motor execution,'' \emph{Neurophotonics}, vol.~5,
  no.~1, p. 011008, 2017.

\bibitem{shibuya2008quantification}
K.~Shibuya, T.~Sadamoto, K.~Sato, M.~Moriyama, and M.~Iwadate, ``Quantification
  of delayed oxygenation in ipsilateral primary motor cortex compared with
  contralateral side during a unimanual dominant-hand motor task using
  near-infrared spectroscopy,'' \emph{Brain research}, vol. 1210, pp. 142--147,
  2008.

\bibitem{Derosiere2014a}
G.~Derosi{\`{e}}re, F.~Alexandre, N.~Bourdillon, K.~Mandrick, T.~E. Ward, and
  S.~Perrey, ``{Similar scaling of contralateral and ipsilateral cortical
  responses during graded unimanual force generation},'' \emph{NeuroImage},
  vol.~85, pp. 471--477, 2014.

\bibitem{wriessnegger2017force}
S.~C. Wriessnegger, D.~Kirchmeyr, G.~Bauernfeind, and G.~R. M{\"u}ller-Putz,
  ``Force related hemodynamic responses during execution and imagery of a hand
  grip task: A functional near infrared spectroscopy study,'' \emph{Brain and
  Cognition}, vol. 117, pp. 108--116, 2017.

\bibitem{friston1994statistical}
K.~J. Friston, A.~P. Holmes, K.~J. Worsley, J.-P. Poline, C.~D. Frith, and
  R.~S. Frackowiak, ``Statistical parametric maps in functional imaging: a
  general linear approach,'' \emph{Human brain mapping}, vol.~2, no.~4, pp.
  189--210, 1994.

\bibitem{ye2009nirs}
J.~C. Ye, S.~Tak, K.~E. Jang, J.~Jung, and J.~Jang, ``{NIRS}-{SPM}: statistical
  parametric mapping for near-infrared spectroscopy,'' \emph{Neuroimage},
  vol.~44, no.~2, pp. 428--447, 2009.

\bibitem{faisal2008noise}
A.~A. Faisal, L.~P. Selen, and D.~M. Wolpert, ``Noise in the nervous system,''
  \emph{Nature reviews neuroscience}, vol.~9, no.~4, p. 292, 2008.

\bibitem{kuboyama2005relationship}
N.~Kuboyama, T.~Nabetani, K.~Shibuya, K.~Machida, and T.~Ogaki, ``Relationship
  between cerebral activity and movement frequency of maximal finger tapping,''
  \emph{Journal of physiological anthropology and applied human science},
  vol.~24, no.~3, pp. 201--208, 2005.

\bibitem{walker2015deep}
I.~Walker, M.~Deisenroth, and A.~Faisal, ``Deep convolutional neural networks
  for brain computer interface using motor imagery,'' \emph{ICL}, p.~68, 2015.

\bibitem{oldfield1971assessment}
R.~C. Oldfield, ``The assessment and analysis of handedness: the edinburgh
  inventory,'' \emph{Neuropsychologia}, vol.~9, no.~1, pp. 97--113, 1971.

\bibitem{cope1988methods}
M.~Cope, D.~Delpy, E.~Reynolds, S.~Wray, J.~Wyatt, and P.~Van~der Zee,
  ``Methods of quantitating cerebral near infrared spectroscopy data,'' in
  \emph{Oxygen Transport to Tissue X}, M.~M. et~al., Ed.\hskip 1em plus 0.5em
  minus 0.4em\relax New York: Springer, 1988, pp. 183--189.

\bibitem{ortega2020hygrip}
P.~Ortega, T.~Zhao, and A.~A. Faisal, ``Hygrip: Full-stack characterisation of
  neurobehavioural signals (fnirs, eeg, emg, force and breathing) during a
  bimanual grip force control task,'' \emph{Frontiers in Neuroscience},
  vol.~14, p. 919, 2020.

\bibitem{kingma2014adam}
D.~P. Kingma and J.~Ba, ``Adam: A method for stochastic optimization,'' in
  \emph{Proceedings of the 2015 3rd International Conference on Learning
  Representations (ICLR)}, 2015, p. n.a.

\bibitem{srivastava2014dropout}
N.~Srivastava, G.~Hinton, A.~Krizhevsky, I.~Sutskever, and R.~Salakhutdinov,
  ``Dropout: a simple way to prevent neural networks from overfitting,''
  \emph{J. Mach. Learning Res.}, vol.~15, no.~1, pp. 1929--1958, 2014.

\bibitem{xiloyannis2015gaussian}
M.~Xiloyannis, C.~Gavriel, A.~A. Thomik, and A.~A. Faisa, ``Gaussian process
  regression for accurate prediction of prosthetic limb movements from the
  natural kinematics of intact limbs,'' in \emph{2015 7th International
  IEEE/EMBS Conference on Neural Engineering (NER)}.\hskip 1em plus 0.5em minus
  0.4em\relax IEEE, 2015, pp. 659--662.

\bibitem{haar2017effector}
S.~Haar, I.~Dinstein, I.~Shelef, and O.~Donchin, ``Effector-invariant movement
  encoding in the human motor system,'' \emph{Journal of Neuroscience},
  vol.~37, no.~37, pp. 9054--9063, 2017.

\bibitem{Ames2019}
K.~C. Ames and M.~M. Churchland, ``Motor cortex signals for each arm are mixed
  across hemispheres and neurons yet partitioned within the population
  response,'' \emph{Elife}, vol.~8, p. e46159, 2019.

\bibitem{gavriel2014comparison}
C.~Gavriel and A.~A. Faisal, ``A comparison of day-long recording stability and
  muscle force prediction between bsn-based mechanomyography and
  electromyography,'' \emph{IEEE Body Sensor Networks (BSN)}, vol.~11, pp.
  69--74, 2014.

\bibitem{fara2014prediction}
S.~Fara, C.~Gavriel, C.~S. Vikram, and A.~A. Faisal, ``Prediction of arm
  end-point force using multi-channel mmg,'' \emph{IEEE Body Sensor Networks
  (BSN)}, vol.~11, pp. 27--32, 2014.

\end{thebibliography}

\end{document}